\documentclass[sigconf,natbib=true,anonymous=false]{acmart}

\copyrightyear{2026}
\acmYear{2026}
\setcopyright{cc}
\setcctype{by}
\acmConference[CIKM '26]{Proceedings of the 35th ACM International Conference on Information and Knowledge Management}{November 07--11, 2026}{Rome, Italy}
\acmBooktitle{Proceedings of the 35th ACM International Conference on Information and Knowledge Management (CIKM '26), November 07--11, 2026, Rome, Italy}
\acmDOI{10.1145/3799682.3841086}
\acmISBN{979-8-4007-2539-5/2026/11}

\usepackage{booktabs}
\usepackage{multirow}
\usepackage{amsmath}
\usepackage{algorithm}
\usepackage{algpseudocode}
\usepackage{xcolor}
\usepackage{url}
\usepackage{booktabs}
\usepackage{pifont}
\usepackage{placeins}
\usepackage{enumitem}
\usepackage{makecell}
\usepackage{subcaption}
\usepackage{float}
\usepackage{threeparttable}
\usepackage{balance}

\newcommand{\cmark}{\ding{51}}%
\newcommand{\xmark}{\ding{55}}%

\title{RAD: Rule-Augmented Relational Anomaly Detection}

\author{Noah Dahle}
\affiliation{%
  \institution{Vanderbilt University}
  \city{Nashville}
  \state{Tennessee}
  \country{USA}
}
\email{noah.dahle@vanderbilt.edu}

\author{Anne Tumlin}
\affiliation{%
  \institution{Vanderbilt University}
  \city{Nashville}
  \state{Tennessee}
  \country{USA}
}
\email{anne.m.tumlin@vanderbilt.edu}

\author{Ngoc Tran}
\affiliation{%
  \institution{Vanderbilt University}
  \city{Nashville}
  \state{Tennessee}
  \country{USA}
}
\email{ngoc.n.tran@vanderbilt.edu}

\author{Xenofon Koutsoukos}
\affiliation{%
  \institution{Vanderbilt University}
  \city{Nashville}
  \state{Tennessee}
  \country{USA}
}
\email{xenofon.koutsoukos@vanderbilt.edu}

\author{Tyler Derr}
\affiliation{%
  \institution{Vanderbilt University}
  \city{Nashville}
  \state{Tennessee}
  \country{USA}
}
\email{tyler.derr@vanderbilt.edu}

\begin{abstract}
Anomaly detection is often applied to data stored in relational databases, yet most existing methods require flattening multiple tables into a single feature matrix. This flattening can obscure entity identity, schema structure, and multi-hop dependencies, limiting the detection of anomalies that depend on relational context rather than isolated feature values. Beyond preserving relational structure, relational anomaly detection raises an additional challenge: how to incorporate symbolic behavioral evidence into learned relational representations. To address these challenges, we study relational anomaly detection, where the goal is to identify anomalous entities or events in a multi-table database. We propose RAD, a rule-augmented relational anomaly detector that combines heterogeneous graph representation learning with refined symbolic rule signals. RAD derives candidate rules from random-forest paths over flattened summaries of the entities or events being scored, refines them into compact interpretable predicates, injects the resulting rule features into the graph model, and learns anomaly scores using reconstruction-based and pairwise-ranking supervision. To evaluate this setting, we introduce a relational anomaly detection benchmark spanning three settings: LANL cybersecurity event detection and two unexpected user-churn anomaly tasks derived from Amazon and H\&M relational databases. Experiments show that RAD improves anomaly ranking over flattened tabular detectors and relational baselines under natural class imbalance, achieving the best average rank on AUROC and AUPRC across the benchmark. Ablations show that direct rule injection and ranking-based supervision are key contributors to performance, while edge reconstruction is not uniformly beneficial. Our code and data are available at: \url{https://github.com/noahd15/RAD_RelationalAnomalyDetection}.

\vspace{-1.5ex}
\end{abstract}

\keywords{relational anomaly detection, symbolic rule mining, relational deep learning, cybersecurity event detection, unexpected user-churn\vspace{-1.25ex}}

\begin{document}

\maketitle

\section{Introduction}
\label{sec:intro}

Anomaly detection is widely used to identify rare or unexpected behavior in domains such as cybersecurity, fraud detection, healthcare, and cyber-physical systems~\cite{anomsurv, deepsurv, coursey2024ft}. 
In many of these settings, the data is not naturally stored as a single table of independent feature vectors. Instead, the data is stored in relational databases containing multiple entity and event types connected through primary-foreign key relationships. For example, a cybersecurity database may include users, machines, and authentication events. A single authentication event may look normal in isolation but become suspicious when linked to an unusual user-machine pair \cite{relbench,fey2024rdl}. This creates two related challenges for anomaly detection on relational databases: preserving relational structure and incorporating symbolic behavioral evidence into learned anomaly representations.

Consider an authentication event that is unremarkable in isolation but anomalous because the user-machine pair has never co-occurred before, the machine is otherwise accessed only by service accounts, and the user authenticated to four other machines in the preceding ten minutes. Recovering these facts requires joins across authentication history and multi-hop neighbor context — information a flattened row-level summary discards.

To apply these methods to relational databases, practitioners often flatten multiple tables into a single feature matrix using joins, aggregations, and hand-crafted features~\cite{relbench,fey2024rdl, ye2025drl}.
Although this makes the data compatible with tabular anomaly detectors, it can also obscure the signals that define relational anomalies. Entity identity, schema roles, multi-hop dependencies, and temporal context may be lost during flattening \cite{relbench,fey2024rdl}. As a result, anomalies whose abnormality depends on cross-table structure can be difficult to detect from flattened row-level features alone.

In this work, we study relational anomaly detection: the task of ranking anomalous target entities or events in a multi-table relational database. In this setting, each target instance must be scored not only from its own attributes but also from its relational context. This context may include linked entities, typed relationships, historical activity, and schema-dependent interactions. Unlike ordinary tabular anomaly detection, relational anomaly detection requires models to preserve and reason over the structure of the database while still producing anomaly scores that are useful under severe class imbalance. Recent work on relational deep learning has shown that relational data can be modeled directly as graphs while preserving table and relationship structure, but this line of work is not itself an anomaly detection framework and does not define anomaly-specific scoring objectives \cite{fey2024rdl,relbench}. Moreover, preserving relational structure alone is not sufficient for relational anomaly detection. Many anomalies correspond to behavioral or semantic patterns that are difficult to capture from relational message passing alone. In practice, anomaly detection systems often rely on symbolic behavioral rules, heuristics, or analyst-defined patterns that encode suspicious combinations of events, entities, or activities. A central challenge is therefore how to incorporate symbolic behavioral evidence into learned relational representations while still preserving end-to-end anomaly-oriented learning.

We propose RAD, a rule-augmented relational anomaly detector for multi-table databases. RAD represents a relational database as a heterogeneous graph, mines symbolic rules from an auxiliary flattened view using random forests \cite{breiman2001random}, injects the resulting rule features into target nodes, and learns anomaly scores via a heterogeneous graph autoencoder with GraphSAGE-style message passing \cite{hamilton2017graphsage}, reconstruction-based scoring, and pairwise ranking supervision. We evaluate RAD on a new relational anomaly detection benchmark spanning LANL cybersecurity event detection and two unexpected user-churn tasks derived from the Amazon and H\&M relational databases in RelBench \cite{relbench}.

Experiments on LANL, Amazon, and H\&M compare RAD against flattened tabular anomaly detectors~\cite{liu2008isolation,OC-SVM,autoencoder,ruff2018deep,ye2025drl}, as well as relational baselines based on Relational Deep Learning~\cite{fey2024rdl}.
More specifically, we structure our empirical evaluation around five research questions: (RQ1) how RAD compares with tabular and relational baselines across relational anomaly tasks; (RQ2) whether mined rule features improve detection; (RQ3) the effect of graph edge reconstruction; (RQ4) how much LLM-assisted rule refinement adds over random-forest rules alone; and (RQ5) how sensitive unexpected-churn results are to anomaly prevalence. Results show that relational modeling and rule augmentation improve anomaly detection, especially on ranking-oriented metrics such as AUPRC and precision@k. These metrics are important because anomaly detection is usually highly imbalanced, and practical systems often care most about whether true anomalies appear near the top of the ranked anomaly list \cite{outdet}. The results also show that rule features are most useful when injected into the graph representation rather than added after graph scoring. Ranking-based supervision further improves anomaly retrieval under class imbalance. Ablations show that direct rule injection and ranking-based supervision are key contributors to performance, while edge reconstruction has dataset-dependent effects and is not uniformly beneficial.

Our contributions are summarized as follows: 

\vspace{-0.5ex}
\begin{itemize}[leftmargin=1.75em]
\item  We formalize relational anomaly detection as distinct from standard tabular and graph anomaly detection, where anomalousness may depend on cross-table, multi-hop, temporal, and schema-dependent interactions.

\item  We propose RAD, a rule-augmented relational anomaly detector for multi-table databases. RAD mines rules from relational summaries, injects rule-derived features into target nodes, and learns anomaly scores with heterogeneous-graph representations, reconstruction, and pairwise-ranking objectives.

\item  We introduce a relational anomaly detection benchmark suite combining LANL cybersecurity event detection with two unexpected user-churn anomaly tasks derived from the Amazon and H\&M datasets~\cite{relbench}. The benchmark provides unified target definitions, anomaly-label constructions, relational graph construction, and ranking-based evaluation protocols.

\item  We empirically compare RAD against tabular anomaly detectors, relational baselines, and targeted ablations, showing that direct rule injection improves anomaly retrieval under severe class imbalance.
\end{itemize}

\vspace{-1ex}
\section{Related Work}
\label{sec:related}

Existing anomaly detection methods differ in how they represent data, how they define anomalousness, and whether they expose interpretable evidence for a detected anomaly. Relational anomaly detection requires all three: the model must preserve multi-table structure, produce useful anomaly rankings under severe class imbalance, and integrate symbolic behavioral evidence that may be difficult to express through relational message passing alone. Prior work addresses these requirements only partially. In this section, we review these existing paradigms and close by positioning our proposed framework against the limitations of current literature.

\vspace{-1ex}
\subsection{From Tabular Anomaly Detection to Relational Structure}

Most classical 
anomaly detection methods assume 
each instance can be represented as an independent feature vector. Isolation Forest~\cite{liu2008isolation}, one-class SVM~\cite{OC-SVM}, autoencoder-based anomaly detection~\cite{autoencoder}, Deep SVDD~\cite{ruff2018deep}, and recent tabular representation-learning methods, e.g., DRL~\cite{ye2025drl}, define anomalousness through isolation, support estimation, reconstruction error, compact latent representations, or learned feature decompositions.

The difficulty is that relational databases are not naturally collections of independent feature vectors. Applying tabular anomaly detectors to multi-table data requires flattening the database through joins, aggregations, and hand-crafted features. Flattening makes standard detectors applicable, but it changes the object being modeled: entity identity, schema roles, multi-hop dependencies, and temporal context must be compressed into row-level summaries. This is a poor fit for anomalies whose suspiciousness is relational, such as an event that appears ordinary in isolation but unusual given the user, machine, product, transaction, or activity history to which it is connected. Thus, tabular anomaly detection provides useful scoring objectives, but its representation is mismatched to relational anomaly detection.

\vspace{-1ex}
\subsection{From Graph Anomaly Detection to Relational Databases}

Graph-based anomaly detection addresses part of this limitation by modeling dependencies among entities. Methods such as DOMINANT~\cite{ding2019deep} and SL-GAD~\cite{zheng2021slgad} use graph neural networks and self-supervised objectives to combine attribute and structural information when detecting anomalies. 
However, graph anomaly detection is usually formulated for graph data rather than relational databases. A relational database contains typed tables, primary-foreign key relationships, event records, entity records, and schema-dependent roles. Treating such data as a generic graph can blur distinctions among entity types and relation types that are important for interpreting anomalies. Relational Deep Learning~\cite{fey2024rdl} and RelBench~\cite{relbench} address this representation issue by preserving multi-table structure and enabling learning directly over relational databases. This makes relational learning a natural foundation for relational anomaly detection.

The remaining gap is the anomaly objective. Relational learning frameworks are typically designed for supervised prediction, and while graph imbalance has been studied across the node, edge, and graph levels~\cite{wang2022imbalanced,zhao2021graphsmote,zhao2021graphsmote}, they are not for ranking rare anomalous target entities under severe class imbalance. In contrast, anomaly detection requires scoring functions and evaluation protocols that prioritize retrieval of rare positives near the top of a ranked list. Our benchmark suite builds on this distinction: LANL provides a cybersecurity relational anomaly setting, while the Amazon and H\&M relational databases from RelBench are adapted into unexpected-churn anomaly detection tasks. The underlying RelBench databases are existing resources, but the anomaly-oriented task construction and unified ranking protocol are specific to this work.

\subsection{From Symbolic Rules to Learned \\Anomaly Scoring}

Symbolic and rule-based methods provide another piece of the problem. Association-rule mining~\cite{agrawal1994fast}, tree-derived rules~\cite{breiman2001random}, and interpretable decision sets~\cite{lakkaraju2016interpretable} can expose human-readable regularities in structured data. Recent work such as ChatRule~\cite{luo2024chatrule} and KnowGraph~\cite{zhou2024knowgraph} further shows how logical rules and knowledge-guided reasoning can capture semantic patterns that are difficult to express through dense embeddings alone.

The limitation is that symbolic rules and learned relational representations are often treated as separate components. Rule-based methods may expose interpretable behavioral patterns, but they are commonly used only for explanation, filtering, or post-hoc reasoning rather than as part of the representation-learning process itself. Conversely, graph and relational neural models can learn expressive embeddings from relational structure, but the resulting anomaly representations may be difficult to connect to explicit behavioral evidence or analyst-defined suspicious patterns. Thus, preserving relational structure alone is often insufficient for relational anomaly detection. A central challenge is how to incorporate symbolic behavioral evidence into learned relational representations while still supporting end-to-end anomaly-oriented learning.

\subsection{Summary of Gaps and Our Contribution}
Positioned against this landscape, RAD's contribution is threefold. First, we formalize relational anomaly detection as a setting distinct from tabular and graph anomaly detection, and introduce a benchmark suite spanning a cybersecurity task with externally verified ground truth and two derived e-commerce tasks under a unified ranking protocol. Second, we identify the point at which symbolic evidence enters the model as a design decision: rule features are injected into target nodes before message passing, allowing them to shape learned representations rather than just adjust scores after the fact, and the ablation in Section~\ref{sec:experiments} evaluates the choice directly against post-hoc fusion. Third, we report empirical findings that hold independent of RAD's specific architecture: edge reconstruction is not uniformly beneficial, and LLM-assisted rule refinement functions well as a compaction and deduplication step rather than as just a source of new signal. With our contributions positioned against this landscape, we next formalize the mathematical foundations of our setting.

\section{Preliminaries}
\label{sec:prelim}

We represent a relational database as a collection of typed tables, 
$\mathcal{D} = \{T_1, T_2, \dots, T_m\}$.
Each table $T_i$ contains rows of a particular entity or event type, and the database schema $\mathcal{S}$ specifies table attributes, primary keys, foreign keys, and relationships among tables\cite{relbench}. For example, rows may correspond to users, computers, authentication events, customers, reviews, transactions, or articles depending on the dataset.

A relational database can be converted to a graph, $\mathcal{G} = (V,E,\tau, \rho)$, 
where each row becomes a node $v \in V$, each primary-foreign key relationship becomes an edge $e \in E$, $\tau(v)$ gives the node type of $v$, and $\rho(e)$ gives the edge type of $e$. This construction follows the relational database-to-graph view used in relational deep learning, where tables and schema relationships define typed nodes and typed edges for message passing \cite{fey2024rdl,relbench}. Unlike a flattened feature matrix, this representation preserves entity identity, table type, and schema-defined relationships.

Let $T_{\mathrm{target}}$ be a designated target table whose rows are the objects to be scored. The corresponding target nodes are denoted $V_{\mathrm{target}} \subset V$. The target type is dataset-dependent. In LANL, target nodes are authentication events, while in the Amazon and H\&M settings, target nodes are users or customers \cite{kent2015cyberdata,relbench}. Each target node $v \in V_{\mathrm{target}}$ may have a binary label $y_v \in \{0,1\}$, where $y_v=1$ indicates an anomaly and $y_v=0$ indicates a normal target node. Labels may be available only for a subset of target nodes.

\section{Problem Definition}
\label{sec:prob_def}
{We formally introduce our proposed problem setting as follows:  
\begin{definition}[Relational Anomaly Detection]
Given a relational database $\mathcal{D}$ with schema $\mathcal{S}$, a designated target table $T_{\mathrm{target}}$, a heterogeneous graph representation $\mathcal{G}=(V,E,\tau)$, target nodes $V_{\mathrm{target}}\subset V$, and anomaly labels for a training subset $V_{\mathrm{train}}\subset V_{\mathrm{target}}$, the goal is to learn an anomaly scoring function,
$s: V_{\mathrm{target}} \rightarrow \mathbb{R}$,  
such that anomalous target nodes receive higher scores than normal target nodes. For labeled target nodes $v_i,v_j \in V_{\mathrm{train}}$ with $y_i=1$ and $y_j=0$, the desired ranking behavior is $s(v_i) > s(v_j)$.

The learned scoring function is evaluated on a held-out test set $V_{\mathrm{test}}\subset V_{\mathrm{target}}$, where
$V_{\mathrm{train}} \cap V_{\mathrm{test}} = \emptyset.$
Evaluation is performed by ranking nodes in $V_{\mathrm{test}}$ according to $s(v)$.
\end{definition}

When timestamps are available, we impose temporal consistency. For a target node $v$ observed at time $t_v$, the scoring function may only use graph structure and node attributes available at or before $t_v$: $s(v) = f(\mathcal{G}_{\leq t_v}, X_{\leq t_v})$. This prevents future information from leaking into the representation of earlier target nodes. The constraint is especially important for event-level cybersecurity detection and churn-style prediction tasks.

Unlike ordinary supervised classification or tabular anomaly detection, the goal here is a ranking that depends on typed relational context in $\mathcal{G}$,
not a calibrated per-instance label.


\vspace{0.5ex}
\section{RAD: Rule-Augmented Relational \\Anomaly Detection}
\label{sec:method}

RAD is a rule-augmented relational anomaly detector for multi-table databases. The model has four stages, shown in Figure~\ref{fig:pipeline}. First, the relational database is converted into a heterogeneous graph that preserves entity types and schema-defined relationships. Second, RAD constructs an auxiliary flattened feature view for the target entities and mines symbolic rules from this view. Third, the selected rules are evaluated on each target entity and injected as binary rule features into the corresponding target nodes. Fourth, a heterogeneous graph autoencoder learns rule-aware relational representations and produces anomaly scores using reconstruction-based and ranking-based objectives.

The design separates rule discovery from relational representation learning. The flattened view is used only to discover interpretable rule predicates. The final anomaly detector operates on the heterogeneous graph, where rules are injected into target nodes before message passing. As a result, symbolic evidence can shape node embeddings and interact with relational context, rather than being added only as a post-hoc adjustment to final anomaly scores.

\vspace{0.5ex}
\subsection{Relational Graph Construction}

Given the heterogeneous graph representation from Section~\ref{sec:prelim}, RAD constructs typed nodes from table rows and typed edges from primary-foreign key relationships. Each target row corresponds to a target node, and neighboring nodes correspond to related entities or events connected through the database schema. The exact target and relation types are dataset-dependent and are described in Section~\ref{sec:bench}.

Each node has an initial feature vector. Since different node types may have different attributes and feature dimensions, RAD uses type-specific input projections before message passing. Graph construction respects the temporal constraint of Section~\ref{sec:prob_def}

The goal of this step is not to flatten the database into one feature matrix, but to preserve typed relational structure so that anomaly scores can depend on both node attributes and relational context.

\begin{figure}[t]
    \centering
    \includegraphics[width=\columnwidth, trim=.5cm 10.25cm 3.825cm 0cm, clip]{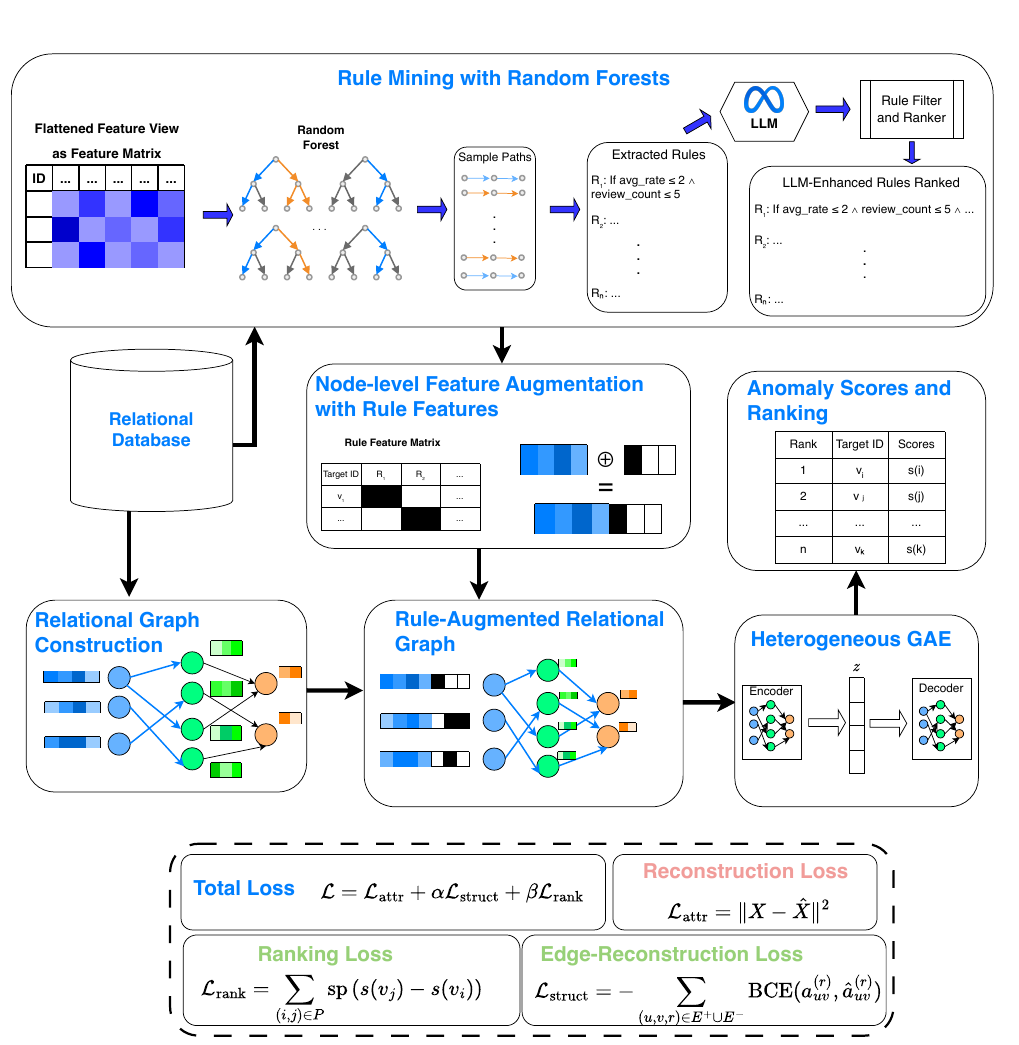}
    \caption{Overview of RAD. A database is represented by a flattened view for rule mining and a heterogeneous graph for representation learning. Random-forest paths are refined into predicates, converted into binary features, and injected into nodes before heterogeneous graph encoding. The model produces anomaly scores and ranks target nodes.}
    \label{fig:pipeline}
\end{figure}

\vspace{0.5ex}
\subsection{Rule Mining and Validation}
\label{sec:rule-mining}

RAD mines symbolic rules from an auxiliary flattened feature view of the target entities. This flattened view is used only for rule discovery, rule evaluation, and tabular baselines; it does not replace the heterogeneous graph used by RAD for relational representation learning. Rule mining's purpose is to extract compact behavioral predicates that can later be injected into target-node features.

Given the flattened training view and the available anomaly labels, RAD trains a random forest classifier after removing identifier, timestamp, label, and unavailable future-information columns. Numeric features are coerced to numeric values, missing values are imputed, and categorical features are encoded before training. Each root-to-leaf path in the trained forest defines a candidate rule. A path is converted into a conjunctive predicate by tracing the feature-threshold decisions from the leaf back to the root. Each split
contributes one condition of the form $x_j \le c$ or $x_j > c$, producing a candidate predicate defined as:
\[
q(v) = c_1(v) \wedge c_2(v) \wedge \cdots \wedge c_m(v).
\]

Following standard rule mining notions of support, confidence, and lift~\cite{agrawal1994fast}, we adapt these quantities to candidate anomaly predicates over target nodes. Let $S$ denote a labeled split, let $v \in S$ be a target node, let $y_v \in \{0,1\}$ be its anomaly label, and let $q(v) \in \{0,1\}$ indicate whether candidate predicate $q$ covers node $v$. The indicator $\mathbf{1}[\cdot]$ equals one when its argument is true and zero otherwise.

The support of predicate $q$ on split $S$ is
\[
\operatorname{support}_{S}(q)
=
\sum_{v \in S} \mathbf{1}[q(v)=1],
\]
which counts the number of nodes covered by the predicate. The confidence of $q$ is the ratio of covered nodes that are anomalous:
\[
\operatorname{confidence}_{S}(q)
=
\frac{
\sum_{v \in S} \mathbf{1}[q(v)=1 \wedge y_v=1]
}{
\sum_{v \in S} \mathbf{1}[q(v)=1]
}.
\]
When the denominator is zero, confidence is defined as zero. 

We also compute lift by comparing predicate confidence against the anomaly prevalence in the same split:
\[
\operatorname{lift}_{S}(q)
=
\frac{\operatorname{confidence}_{S}(q)}{\pi_S},
\qquad
\pi_S =
\frac{1}{|S|}
\sum_{v \in S} \mathbf{1}[y_v=1].
\]
Lift measures whether a predicate covers anomalies at a higher rate than expected from the base anomaly prevalence.

Candidate predicates are filtered before being used as features. Predicates that reference identifiers, label fields, unavailable columns, or future information are removed. Predicates that fail parsing, fail execution, or do not satisfy the minimum coverage requirements are also discarded. The remaining predicates are ranked using support, confidence, and lift computed only from the training or validation data designated for rule selection. Duplicate predicates are removed.

From the retained candidate predicates, RAD samples a smaller set of seed rules for optional LLM-assisted refinement. The LLM is used only as a constrained rule-proposal mechanism. Given the flattened feature
schema, feature statistics, compact positive examples from the training data, and sampled random-forest seed rules, it proposes SQL \texttt{WHERE} predicates that prune, tighten, combine, or generalize the seed predicates. The LLM does not assign anomaly labels, produce anomaly scores, introduce external attributes, or access test labels.

Generated predicates are validated before use. Each generated predicate is executed against the flattened rule-selection view and rescored using the same support, confidence, and lift definitions above. Invalid predicates and predicates that fail the required coverage checks are discarded. The highest-ranked valid predicates are retained as the final rule set.

The final rule set is frozen after training and validation. At test time, RAD only evaluates these fixed predicates on held-out target entities. No rule mining, rule selection, threshold relaxation, or predicate updating uses test labels.

Each retained predicate is converted into a binary rule feature. For rule $k$, the rule feature for target entity $v$ is
\[
r_k(v) =
\begin{cases}
1, & \text{if } v \text{ satisfies rule } k,\\
0, & \text{otherwise}.
\end{cases}
\]
The full rule vector is
\[
r(v) = [r_1(v), r_2(v), \ldots, r_K(v)].
\]
The resulting binary rule matrix is aligned with the corresponding target nodes and injected into the heterogeneous graph as described
in Section~\ref{sec:rule-augmented-features}.

\subsection{Rule-Augmented Node Features}
\label{sec:rule-augmented-features}

After rule mining and validation, RAD evaluates the final rule set on each target entity. This produces a binary rule vector $r(v)$ for each target node $v \in V_\text{target}$. Each entry indicates whether the target entity satisfies one selected rule.

RAD injects this rule vector directly into the target node features before message passing. For each target node $v$, the augmented feature vector is
\[
\tilde{x}_v=\left[x_v \| r(v)\right],
\]
where $x_v$ is the original target-node feature vector and $\|$ denotes concatenation. Non-target nodes keep their original feature vectors.

This direct injection is a key design choice. Because rule features are added before graph encoding, rule evidence can influence message passing and representation learning. A target node's embedding can therefore reflect not only its own attributes and relational neighborhood, but also the symbolic rules it satisfies. This differs from post-hoc rule fusion, where rule features are added after the graph model has already produced anomaly scores. Post-hoc fusion can adjust final scores, but it cannot change the representations learned by the graph encoder. We evaluate this design choice in the direct-injection versus post-hoc-fusion ablation in Section~\ref{sec:results}.

\subsection{Heterogeneous Graph Encoder}
\label{sec:heterogeneous-encoder}

RAD uses a heterogeneous graph encoder to learn rule-aware node
representations from the augmented relational graph. Because
different node types may have different feature spaces, each node type
first uses a type-specific input projection. For target nodes, the input
is the rule-augmented feature vector $\tilde{x}_v$; for non-target
nodes, the input is the original feature vector $x_v$.

After projection, RAD applies relation-aware message passing over the
typed edges of the relational graph. Messages are aggregated separately
for each relation type and then combined to update the node
representation. This preserves schema information by allowing messages
arriving through different database relationships to contribute
separately before being merged.

In our implementation, the encoder is a two-layer heterogeneous
GraphSAGE model with mean aggregation~\cite{hamilton2017graphsage}.
After the final message-passing layer, each target node $v$ has an
embedding $z_v$ that encodes its original attributes, injected rule
features, and typed relational neighborhood. These embeddings are used
by the reconstruction and ranking objectives described next.

Formally, for node $v$ of type $\tau(v)$, the encoder computes
\begin{align*}
h_v^{(0)} &= W_{\tau(v)}\,\tilde{x}_v + b_{\tau(v)}, \\
m_v^{(l,r)} &= \frac{1}{|\mathcal{N}_r(v)|}\sum_{u \in \mathcal{N}_r(v)} h_u^{(l-1)}, \\
h_v^{(l)} &= \sigma\!\left(W^{(l)}_{\text{self}} h_v^{(l-1)} + \sum_{r \in \mathcal{R}(v)} W^{(l)}_r\, m_v^{(l,r)}\right), \\
z_v &= h_v^{(L)}, \qquad L = 2,
\end{align*}
where $\mathcal{N}_r(v)$ denotes the neighbors of $v$ under relation $r$, $\mathcal{R}(v)$ the relation types incident to $\tau(v)$, and $\tilde{x}_v$ the rule-augmented feature vector from Section~4.3 for target nodes and $x_v$ otherwise. 
We use GraphSAGE-style mean aggregation because it is inductive (required for scoring LANL's sampled subgraphs at test time), stable under variable-degree sampled neighborhoods, and matches the RDL baseline encoder in Section~\ref{sec:baselines}, isolating the effect of rule injection from backbone choice.

\subsection{Anomaly Scoring and Training Objective}
RAD scores target nodes using a graph autoencoder objective with optional structure reconstruction and ranking-based anomaly supervision, following the general use of graph autoencoders to learn latent node representations and reconstruct graph structure~\cite{kipf2016vgae}.

\subsubsection{Attribute Reconstruction}

The attribute decoder attempts to reconstruct the rule-augmented target-node features $\tilde{x}_v$ from the target-node embedding $z_v$, which we formalize as:
\[
\hat{\tilde{x}}_v = g_\phi(z_v).
\]

The attribute reconstruction loss for target node $v$ is
\[
\mathcal{L}_\text{attr}(v)
=
\left\|
\hat{\tilde{x}}_v-\tilde{x}_v
\right\|_2^2.
\]

This term encourages the encoder to learn representations that preserve target-node attributes and rule features. The reconstruction error also provides an anomaly signal. Target nodes that do not follow common attribute, rule, or relational patterns should be harder to reconstruct.

\subsubsection{Edge Reconstruction}

RAD can also include an edge reconstruction term. For an edge $(u,v,r)$ of relation type $r$, the structure decoder predicts whether the edge exists from the node embeddings. This can be formalized as follows:
\[
\hat{a}^{(r)}_{uv}
=
\sigma\left(z^\top_u W_r z_v\right),
\]
where $W_r$ is a relation-specific decoder parameter and $\hat{a}^{(r)}_{uv}$ is the predicted probability that an edge of type $r$ exists between nodes $u$ and $v$ in the relational graph.

The edge reconstruction loss is binary cross-entropy over observed edges and sampled non-edges:
\[
\mathcal{L}_{\mathrm{struct}}
=
-\sum_{(u,v,r)\in E^+ \cup E^-}
\left[
a^{(r)}_{uv}\log \hat{a}^{(r)}_{uv}
+
\left(1-a^{(r)}_{uv}\right)\log
\left(1-\hat{a}^{(r)}_{uv}\right)
\right].
\]
Here, $E^+$ is the set of observed typed edges, $E^-$ is a set of sampled negative edges, and $a^{(r)}_{uv} \in \{0,1\}$ indicates whether the typed edge exists.

We treat edge reconstruction as optional because it may help when anomalousness is expressed through unusual links, but it may also overemphasize sampled graph topology. We therefore evaluate both BCE and no-BCE variants in Section~\ref{sec:results}. The no-BCE variant sets the edge-reconstruction weight to zero.

\subsubsection{Ranking-Based Anomaly Supervision}

Reconstruction alone is an indirect anomaly detection objective. A powerful autoencoder may reconstruct some anomalous nodes well, and the practical goal is to rank rare anomalies near the top rather than to produce calibrated binary predictions. For this reason, RAD uses pairwise ranking supervision when anomaly labels are available, following pairwise learning-to-rank objectives that encourage positive instances to score above negative instances~\cite{burges2005ranknet}.

Let $P$ be the set of labeled anomalous target nodes and $N$ be the set of presumed-normal target nodes. RAD samples pairs $(i,j) \sim P\times N$ and encourages anomalous nodes to receive higher anomaly scores than normal nodes:
\[
\mathcal{L}_\text{rank}
=
\mathbb{E}_{(i,j)\sim P\times N}
\left[
\operatorname{softplus}(s_j-s_i)
\right].
\]
Here, $s_i$ and $s_j$ are anomaly scores for the anomalous and normal target nodes, respectively. In our implementation, the anomaly score is based on reconstruction error, optionally combined with the edge-reconstruction score when used.

The full training objective is
\[
\mathcal{L}
=
\mathcal{L}_\text{attr}
+
\alpha\mathcal{L}_\text{struct}
+
\beta\mathcal{L}_\text{rank},
\]
where $\alpha$ controls the contribution of edge reconstruction and $\beta$ controls the contribution of ranking supervision. In no-BCE ablations, $\alpha=0$. This objective aligns RAD with anomaly ranking under class imbalance while still using reconstruction to learn rule-aware relational representations.
\section{Relational Anomaly Detection Benchmark}
\label{sec:bench}

We introduce a relational anomaly detection benchmark suite for evaluating anomaly ranking in multi-table databases. The suite contains one cybersecurity task based on the LANL Comprehensive, Multi-Source Cybersecurity Events dataset~\cite{kent2015cyberdata} and two e-commerce tasks derived from the Amazon and H\&M relational databases in RelBench~\cite{relbench}. The underlying Amazon and H\&M databases are not new; our contribution is to adapt them into unexpected-churn anomaly detection tasks and evaluate all three tasks under a unified relational anomaly detection protocol.

\subsection{Benchmark Design Principles}
\label{sec:benchmark-design}

A task is suitable for relational anomaly detection when anomalousness may depend on relational context rather than only on the target entity's own attributes. We therefore select tasks that contain multiple schema-linked tables, define a target entity or event type, provide anomaly labels or a principled anomaly-label construction, contain rare positives suitable for ranking-based evaluation, and include temporal information when needed to prevent leakage. These criteria distinguish relational anomaly detection from flattened tabular anomaly detection: the target score may depend on linked entities, typed relationships, historical behavior, and multi-hop schema-dependent context.

Table~\ref{tab:benchmark-summary} summarizes the benchmark tasks, target entities, database scale, and anomaly definitions. For Amazon and H\&M, the default unexpected-churn anomaly rate is 1.5\%, with 0.5\% and 3.0\% used in the prevalence-sensitivity sweep.

\begin{table}[t]
\centering
\setlength{\tabcolsep}{2pt}
\caption{Summary of the relational anomaly detection benchmark tasks. LANL is treated as a multi-table event database, while Amazon and H\&M are derived from relational user-behavior databases and converted into unexpected churn anomaly detection tasks.}
\vspace{-1ex}
\label{tab:benchmark-summary}
\small
\resizebox{0.9\columnwidth}{!}{%
\begin{tabular}{lcccccl}
\toprule
Task & Domain & Target & Tables & Records & Targets & Anomaly \\
\midrule
LANL & Cybersecurity &
Auth event &
7 &
1.65B &
1.64B &
Red-team auth. \\

Amazon & E-commerce &
Customer &
3 &
15.0M &
1.85M &
\begin{tabular}[c]{@{}l@{}}Unexpected\\review churn\end{tabular} \\

H\&M & Retail &
Customer &
3 &
16.7M &
1.37M &
\begin{tabular}[c]{@{}l@{}}Unexpected\\purchase churn\end{tabular} \\
\bottomrule
\end{tabular}%
}
\vspace{-2ex}
\end{table}

\subsection{LANL Cybersecurity Event Detection}
\label{sec:lanl}

The LANL cybersecurity dataset contains enterprise authentication events, process execution logs, DNS activity, network flows, and red-team attack labels~\cite{kent2015cyberdata}. We use authentication events as target entities and red-team authentication events as anomalies. This task represents event-level anomaly detection in a relational cybersecurity database.
LANL is well suited for relational anomaly detection because suspicious authentication behavior often depends on context: a login may appear normal as a row but become suspicious when linked to unusual users, machines, process activity, etc. 
As the full dataset is large,  
we evaluate sampled relational subgraphs centered on authentication events. Anchor events are sampled across the full time range, with a small number of red-team events included to ensure positives. For each anchor, we retrieve linked users, source and destination computers, and nearby process, DNS, and flows when available.

\subsection{Unexpected User-Churn Detection}
\label{sec:unexpected-churn}

We derive two unexpected-churn tasks from relational user-behavior databases. Unlike ordinary churn prediction, these tasks treat churn as anomalous only when it occurs without a preceding activity decline that would make the churn expected. This distinction is important since 
predictable disengagement is not necessarily anomalous. The resulting tasks ask whether relational context helps identify users whose behavior changes in surprising ways across linked reviews, purchases, products, transactions, and item metadata.

\subsubsection{Amazon Review Churn}
\label{sec:amazon-churn}

The Amazon task scores customers whose reviewing behavior ends unexpectedly. Each customer is connected to review events and reviewed products, allowing the relational graph to represent not only how often a customer reviews, but also what kinds of products they review and how their activity relates to product-level context. The auxiliary flattened view for Amazon summarizes each customer's own review activity, review counts, rating history, product diversity as a count, and review-history span, and does not incorporate attributes of the reviewed products themselves. Mined rules for this task are therefore restricted to one-hop customer-level aggregates; multi-hop product context reaches the model only through message passing over the customer-review-product graph, not through rule features. 

\subsubsection{H\&M Purchase Churn}
\label{sec:hm-churn}

The H\&M task evaluates unexpected disappearance from purchase activity. Customers are linked to transaction records and articles, with article attributes providing item-level context for purchase behavior. This task differs from Amazon because activity is expressed through purchases rather than reviews, and because product metadata can help distinguish ordinary inactivity from unusual changes in shopping behavior. The flattened view summarizes recent transactions, price history, article diversity, and transaction-history span, while the relational graph preserves customer-transaction-article links. 

\subsubsection{Unexpected-Churn Label Construction}
\label{sec:unexpected-churn-labels}

For Amazon and H\&M, we convert churn labels into unexpected-churn anomaly labels by removing churn cases that are explained by a strong prior activity decline.

A user is therefore anomalous only if they churn and their previous activity does not already indicate an expected disengagement pattern.
The unexpected-churn label is derived from the original RelBench \cite{relbench} churn label via the activity-decline filter below, not an independent construction. The filter removes only those churned users whose own prior activity history already signals disengagement; it does not introduce a new labeling criterion unrelated to the underlying data. The prevalence sweep in Section~\ref{sec:experiments} varies the filtering threshold across $0.5\%, 1.5\%,$ and $3.0\%$ and shows that RAD's ranking advantage over the supervised RDL baseline is stable across this range, which is the evidence that the default $1.5\%$ threshold was not selected to favor a particular result. 

Let $y^{\mathrm{churn}}_u \in \{0,1\}$ denote the original churn label for user $u$. We compute a recent activity-change score $g_u$ and a baseline activity level $\mu_u$ from the user's activity history before the prediction time. A user is marked as having an expected activity decline when
\[
D(u) = \mathbf{1}\left[g_u \le -\theta(\mu_u)\right],
\]
where
\[
\theta(\mu_u) = \lambda_0 + \log_{\lambda_1}(\mu_u+1)(1+\lambda_2\mu_u).
\]
The threshold adapts to the user's baseline activity level, so users with different activity histories are not judged using the same fixed decline cutoff. In our experiments, we use $\lambda_0 = 0.05$, $\lambda_1 = 10.0$, and $\lambda_2 = 0.02$.

The filtered anomaly label is
\[
y_u^{\mathrm{anom}} = y_u^{\mathrm{churn}}(1-D(u)).
\]
Thus, a user is labeled anomalous only if the user churned and was not already identified as having an expected activity decline. Churned users with an expected prior decline are removed from the positive anomaly class. After filtering, training positives are downsampled to a target anomaly prevalence of 1.5\%, and the validation set is downsampled to match the resulting training prevalence. Evaluation is performed on held-out target users using ranking metrics.

\begin{figure}[t]
    \centering
    \includegraphics[width=\linewidth]{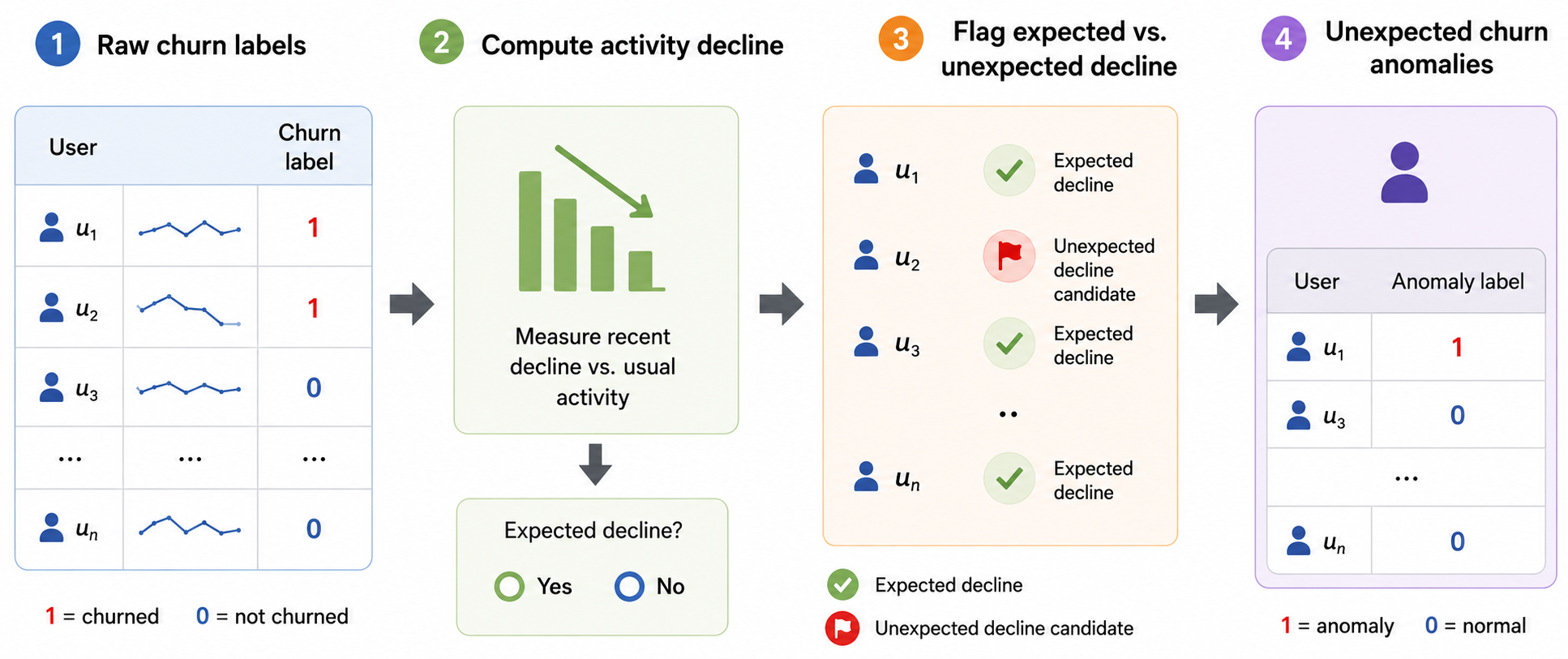}
    \Description{Unexpected-churn filtering keeps churned users without a prior activity decline and removes churned users with an expected decline.}
    \vspace{-2ex}
    \caption{Unexpected-churn filtering. Churned users with an expected prior activity decline are removed; remaining churned users are treated as unexpected-churn anomalies.}
    \label{fig:churn-filter}
        \vspace{-3ex}
\end{figure}

\subsubsection{Prevalence Sweep for Label Sensitivity}
\label{sec:prevalence-sweep}

The unexpected-churn label depends on the strictness of the filtering threshold used to remove expected churn cases. To test whether the benchmark is robust to this design choice, we perform a prevalence sweep over the target anomaly labeling rate. After constructing candidate unexpected-churn positives, we vary the retained positive rate to produce target prevalences of 0.5\%, 1.5\%, and 3.0\%. The 1.5\% task is used as the default benchmark configuration, while the 0.5\% and 3.0\% tasks evaluate sensitivity under stricter and more permissive anomaly definitions.

For each prevalence level, we preserve the same train/validation/test protocol and compare RAD against the supervised RDL baseline using AUROC and AUPRC. The sweep changes only the anomaly-label filtering rate; the relational graph construction, feature views, model architecture, and evaluation protocol are otherwise held fixed. This isolates the effect of label prevalence from changes in model design.

\subsection{Dataset-Specific Flattened Views and \\Rule Mining}
\label{sec:dataset-specific-rule-mining}

Although RAD operates on heterogeneous relational graphs, rule mining and tabular baselines require an auxiliary flattened view for each target entity. Table~\ref{tab:feature-views-rf-settings} summarizes the dataset-specific flattened feature views along with random-forest and predicate-filtering settings that instantiate Section~\ref{sec:rule-mining}'s  generic rule-mining procedure.

\begin{table}[t]
\centering
\caption{Feature views and random-forest rule-mining settings by dataset.}
\vspace{-1.5ex}
\label{tab:feature-views-rf-settings}
\footnotesize
\resizebox{\columnwidth}{!}{%
\begin{tabular}{lccc}
\toprule
Setting & LANL & Amazon & H\&M \\
\midrule
Feature groups &
\begin{tabular}[t]{@{}c@{}}Auth attrs.; user-pair \\ novelty; recent auth., \\ DNS, flow, process\end{tabular} &
\begin{tabular}[t]{@{}c@{}}Recent reviews; \\ rating history; \\ product diversity; \\ review-history span\end{tabular} &
\begin{tabular}[t]{@{}c@{}}Recent purchases; \\ price history; article \\ diversity; activity \\ trend/level\end{tabular} \\
\addlinespace
Training cap & 500K & 200K & 200K \\
Forest size & 1200 & 200 & 200 \\
Depth / leaf & None / 1 & 10 / 5 & 10 / 5 \\
Class weight & Balanced & Balanced & Balanced \\
Path extraction & First 200 & First 100 & First 100 \\
Leaf filter & $\geq$8, $p_+ \geq .50$ & $\geq$5, $p_+ \geq .50$ & $\geq$5, $p_+ \geq .50$ \\
Selection split & Validation & Train sample & Train sample \\
Rule filter & Supp. $\geq$3, conf. $\geq$.02 & Supp. $\geq$1 & Supp. $\geq$1 \\
Seed / pool & 30 / 2000 & 30 / 2000 & 30 / 2000 \\
\bottomrule
\end{tabular}%
}
\vskip -2ex
\end{table}
\section{Experiments}
\label{sec:experiments}

The experiments evaluate whether RAD improves relational anomaly detection through relational representation learning, rule augmentation, and anomaly-oriented supervision, returning to the five research questions introduced in Section 1.

\subsection{Experimental Protocol}

We evaluate RAD on the benchmark from Section~\ref{sec:bench}, using the pipeline described in Section~\ref{sec:method}. Each setting defines a target entity type (authentication events for LANL, customers for Amazon and H\&M); the model produces a continuous anomaly score per target node, evaluated only on held-out entities. Flattened rows are aligned with graph target nodes by target identifier so all methods share the same target population and splits. The supervised MLP uses ground-truth training labels and is evaluated on the same held-out test entities; test labels are used only for evaluation. Results are reported over ten random seeds as mean ± standard deviation, using AUROC, AUPRC, and precision@k, with AUPRC and precision@k emphasized because anomaly detection is highly imbalanced and practical systems prioritize retrieving true anomalies near the top of a ranked list \cite{outdet}.

\subsection{Implementation Details}

RAD uses a two-layer relation-aware GraphSAGE encoder with mean aggregation, hidden dimension 128, Adam optimization, learning rate 0.001, MSE attribute reconstruction, optional BCE edge reconstruction, and early stopping on validation AUPRC. The structure weight $\alpha$, ranking weight $\beta$, and model hyperparameters are selected using Optuna on the validation split, with validation AUPRC as the optimization objective. The selected configuration is then fixed for held-out test evaluation. 

The supervised MLP operates on the flattened target-entity features and is trained using binary cross-entropy on ground-truth training labels. For LANL, MLP hyperparameters are selected using the validation split. For Amazon and H\&M, the MLP uses the default, untuned hyperparameters. In all cases, the resulting model is evaluated on the held-out test split over ten random seeds.

For experiments with edge reconstruction, the structure reconstruction weight $\alpha$ controls the contribution of the binary cross-entropy edge reconstruction term; no-BCE ablations set $\alpha$ to zero.

For LANL, rule mining uses the validation-based filtering and support thresholds described in Table~\ref{tab:feature-views-rf-settings}.

\subsection{Baselines}
\label{sec:baselines}
We compare against flattened tabular anomaly detectors, including Isolation Forest~\cite{liu2008isolation}, Deep SVDD~\cite{ruff2018deep}, autoencoder reconstruction~\cite{autoencoder}, AE+Rank, and DRL~\cite{ye2025drl}. We also compare against relational deep learning baselines that operate directly on the heterogeneous graph without rule augmentation~\cite{fey2024rdl}. Finally, we include a supervised MLP trained on the flattened target-entity features as a fully supervised reference baseline. This provides a reference for how the anomaly detection methods perform relative to a model trained directly using ground-truth anomaly labels. These baselines separate the effects of tabular scoring, relational representation learning, rule augmentation, ranking supervision, and full label supervision. Table~\ref{tab:baseline-summary} summarizes the baselines.

\begin{table}[h]
\centering
\caption{\small Baselines compared in our experiments, categorized by input representation and supervision signal.}
\label{tab:baseline-summary}
\small
\begin{threeparttable}
\begin{tabular}{lcccc}
\toprule
Model & Tab. & Graph & Rank & Sup. \\
\midrule
Isolation Forest~\cite{liu2008isolation}
& \cmark & \xmark & \xmark & \xmark \\
Deep SVDD~\cite{ruff2018deep}
& \cmark & \xmark & \xmark & \xmark \\
AE~\cite{autoencoder}
& \cmark & \xmark & \xmark & \xmark \\
AE + Rank
& \cmark & \xmark & \cmark & \xmark \\
DRL~\cite{ye2025drl}
& \cmark & \xmark & \xmark & \xmark \\
RDL GNN~\cite{fey2024rdl}
& \xmark & \cmark & \xmark & \xmark \\
Supervised MLP
& \cmark & \xmark & \xmark & \cmark \\
\bottomrule
\end{tabular}
\begin{tablenotes}
\footnotesize
\item ``Tab.'' indicates flattened tabular input, ``Graph'' indicates heterogeneous relational graph input, ``Rank'' indicates pairwise anomaly-ranking supervision, and ``Sup.'' indicates direct supervision using ground-truth anomaly labels.
\end{tablenotes}
\end{threeparttable}
\end{table}

\begin{table*}[!t]
\centering
\scriptsize
\caption{Main results across datasets over 10 seeds. Results are reported as mean $\pm$ standard deviation. Avg. rank columns report the average rank per metric across datasets; lower is better.}
\label{tab:main-results}

\begin{tabular}{lcccccccccccc}
\toprule
& \multicolumn{3}{c}{LANL}
& \multicolumn{3}{c}{Amazon}
& \multicolumn{3}{c}{H\&M}
& \multicolumn{3}{c}{Avg. Rank} \\
\cmidrule(lr){2-4}
\cmidrule(lr){5-7}
\cmidrule(lr){8-10}
\cmidrule(lr){11-13}

Model
& AUROC & AUPRC & P@k
& AUROC & AUPRC & P@k
& AUROC & AUPRC & P@k
& AUROC & AUPRC & P@k \\
\midrule

RAD (rules, no BCE)
& $\mathbf{0.996}$ {\tiny $\pm 0.001$}
& $\mathbf{0.659}$ {\tiny $\pm 0.053$}
& $\mathbf{0.601}$ {\tiny $\pm 0.012$}
& $0.693$ {\tiny $\pm 0.010$}
& $\mathbf{0.043}$ {\tiny $\pm 0.003$}
& $0.050$ {\tiny $\pm 0.007$}
& $\mathbf{0.828}$ {\tiny $\pm 0.071$}
& $\mathbf{0.143}$ {\tiny $\pm 0.017$}
& $\mathbf{0.129}$ {\tiny $\pm 0.022$}
& $\mathbf{2.0}$ & $\mathbf{1.2}$ & $\mathbf{2.0}$ \\

RAD (rules, BCE)
& $0.860$ {\tiny $\pm 0.155$}
& $\mathit{0.557}$ {\tiny $\pm 0.029$}
& $\mathit{0.572}$ {\tiny $\pm 0.008$}
& $0.695$ {\tiny $\pm 0.010$}
& $\mathbf{0.043}$ {\tiny $\pm 0.004$}
& $\mathit{0.051}$ {\tiny $\pm 0.004$}
& $\mathit{0.787}$ {\tiny $\pm 0.074$}
& $\mathit{0.110}$ {\tiny $\pm 0.020$}
& $\mathit{0.125}$ {\tiny $\pm 0.022$}
& $\mathit{3.3}$ & $\mathit{1.8}$ & $\mathit{2.3}$ \\

RDL weighted~\cite{fey2024rdl}
& $0.553$ {\tiny $\pm 0.158$}
& $0.007$ {\tiny $\pm 0.014$}
& $0.021$ {\tiny $\pm 0.064$}
& $\mathbf{0.720}$ {\tiny $\pm 0.002$}
& $\mathit{0.037}$ {\tiny $\pm 0.000$}
& $\mathbf{0.062}$ {\tiny $\pm 0.004$}
& $0.700$ {\tiny $\pm 0.005$}
& $0.035$ {\tiny $\pm 0.001$}
& $\mathit{0.064}$ {\tiny $\pm 0.013$}
& $4.7$ & $4.8$ & $3.7$ \\

RDL unweighted~\cite{fey2024rdl}
& $0.685$ {\tiny $\pm 0.205$}
& $0.015$ {\tiny $\pm 0.028$}
& $0.026$ {\tiny $\pm 0.078$}
& $\mathit{0.713}$ {\tiny $\pm 0.002$}
& $\mathit{0.037}$ {\tiny $\pm 0.001$}
& $\mathbf{0.062}$ {\tiny $\pm 0.005$}
& $0.693$ {\tiny $\pm 0.019$}
& $0.039$ {\tiny $\pm 0.004$}
& $\mathit{0.064}$ {\tiny $\pm 0.016$}
& $4.3$ & $3.8$ & $3.3$ \\

AE+Rank~\cite{autoencoder}
& $0.943$ {\tiny $\pm 0.035$}
& $0.525$ {\tiny $\pm 0.078$}
& $0.540$ {\tiny $\pm 0.065$}
& $0.518$ {\tiny $\pm 0.014$}
& $0.016$ {\tiny $\pm 0.001$}
& $0.015$ {\tiny $\pm 0.003$}
& $0.456$ {\tiny $\pm 0.024$}
& $0.013$ {\tiny $\pm 0.000$}
& $0.005$ {\tiny $\pm 0.004$}
& $5.2$ & $5.2$ & $5.7$ \\

DRL~\cite{ye2025drl}
& $0.673$ {\tiny $\pm 0.159$}
& $0.004$ {\tiny $\pm 0.002$}
& $0.005$ {\tiny $\pm 0.004$}
& $0.463$ {\tiny $\pm 0.022$}
& $0.013$ {\tiny $\pm 0.001$}
& $0.003$ {\tiny $\pm 0.000$}
& $0.456$ {\tiny $\pm 0.012$}
& $0.013$ {\tiny $\pm 0.001$}
& $0.009$ {\tiny $\pm 0.009$}
& $7.5$ & $8.2$ & $7.2$ \\

AE~\cite{autoencoder}
& $0.888$ {\tiny $\pm 0.046$}
& $0.009$ {\tiny $\pm 0.003$}
& $0.000$ {\tiny $\pm 0.000$}
& $0.474$ {\tiny $\pm 0.013$}
& $0.014$ {\tiny $\pm 0.001$}
& $0.014$ {\tiny $\pm 0.009$}
& $0.408$ {\tiny $\pm 0.020$}
& $0.012$ {\tiny $\pm 0.000$}
& $0.007$ {\tiny $\pm 0.005$}
& $6.7$ & $7.3$ & $7.8$ \\

Deep SVDD~\cite{ruff2018deep}
& $0.736$ {\tiny $\pm 0.073$}
& $0.005$ {\tiny $\pm 0.002$}
& $0.001$ {\tiny $\pm 0.003$}
& $0.458$ {\tiny $\pm 0.031$}
& $0.013$ {\tiny $\pm 0.001$}
& $0.003$ {\tiny $\pm 0.001$}
& $0.412$ {\tiny $\pm 0.023$}
& $0.012$ {\tiny $\pm 0.001$}
& $0.002$ {\tiny $\pm 0.003$}
& $7.7$ & $8.7$ & $8.8$ \\

Isolation Forest~\cite{liu2008isolation}
& $0.644$ {\tiny $\pm 0.018$}
& $0.003$ {\tiny $\pm 0.000$}
& $0.000$ {\tiny $\pm 0.000$}
& $0.450$ {\tiny $\pm 0.002$}
& $0.013$ {\tiny $\pm 0.000$}
& $0.001$ {\tiny $\pm 0.000$}
& $0.402$ {\tiny $\pm 0.014$}
& $0.012$ {\tiny $\pm 0.000$}
& $0.004$ {\tiny $\pm 0.005$}
& $9.7$ & $9.3$ & $9.5$ \\

Supervised MLP
& $\mathit{0.965}$ {\tiny $\pm 0.003$}
& $0.050$ {\tiny $\pm 0.011$}
& $0.093$ {\tiny $\pm 0.035$}
& $0.659$ {\tiny $\pm 0.024$}
& $0.027$ {\tiny $\pm 0.003$}
& $0.041$ {\tiny $\pm 0.009$}
& $0.642$ {\tiny $\pm 0.028$}
& $0.027$ {\tiny $\pm 0.005$}
& $0.039$ {\tiny $\pm 0.023$}
& $4.0$ & $4.7$ & $4.7$ \\

\bottomrule
\end{tabular}

\vspace{3ex}
\end{table*}

\subsection{Evaluation Metrics}

All methods produce a continuous anomaly score for each held-out target entity, where larger scores indicate greater anomalousness. We report AUROC for global ranking quality, AUPRC for precision-recall performance on the anomalous class, and precision@k for the fraction of true anomalies among the top-$k$ ranked targets, with $k$ set to the number of positives in the test set. We emphasize AUPRC and precision@k because precision-recall evaluation is more informative than ROC analysis under severe class imbalance, and because practical anomaly detection systems often inspect only the top-ranked alerts~\cite{davis2006precision,outdet}.

AUROC is the probability that a uniformly drawn positive target node receives a higher score than a uniformly drawn negative one. AUPRC is the area under the precision-recall curve, summed across recall levels. Precision@k is the fraction of true positives among the top-$k$ ranked targets, with $k$ set to the number of positives in the evaluated split.

\section{Results}
\label{sec:results}

The results evaluate whether RAD improves relational anomaly detection through relational graph modeling, rule augmentation, and anomaly-oriented supervision. We focus primarily on AUPRC and precision@k because the benchmark tasks are highly imbalanced and require retrieving rare anomalies near the top of a ranked list.

\subsection{RQ1: Overall Relational Anomaly Detection Performance}
\label{sec:rq1-results}

Table~\ref{tab:main-results} reports the main results across LANL, Amazon, and H\&M. RAD achieves the strongest average AUPRC rank across the benchmark, with the clearest gains on LANL. On LANL, RAD substantially outperforms flattened tabular detectors and relational baselines on AUPRC and precision@k, indicating that relational structure and rule augmentation are especially useful for retrieving rare suspicious authentication events.

On Amazon, the supervised RDL baselines achieve stronger AUROC and precision@k than RAD, but RAD achieves higher AUPRC. This suggests that RDL provides strong global ranking on this task, while RAD improves precision-recall behavior over the anomalous class. On H\&M, RAD achieves the best performance across AUROC, AUPRC, and precision@k. Overall, these results support the central motivation of relational anomaly detection: anomaly ranking benefits from modeling both relational context and rule-derived behavioral evidence, especially under severe class imbalance.

\vspace{0.5ex}
\subsection{RQ2: Effect of Rule Augmentation}
\label{sec:rq2-results}

Figure~\ref{fig:ablation} evaluates the contribution of rule augmentation and rule construction choices. Rule-augmented RAD variants generally improve anomaly retrieval relative to variants without rule features, especially on AUPRC and precision@k. This indicates that symbolic predicates provide useful behavioral evidence beyond the original target-node attributes and relational graph structure.

The comparison between rule-construction variants shows that compact predicate-based rules are more useful than raw decision-feature injection. Random-forest path rules already provide meaningful anomaly signal, while LLM-assisted refinement helps filter, tighten, and compact the candidate predicates before they are injected into the graph. These results support the design choice of converting mined rules into explicit binary rule features rather than treating the random forest only as a black-box auxiliary model.

\begin{figure}[t]
    \centering
    \includegraphics[width=\linewidth]{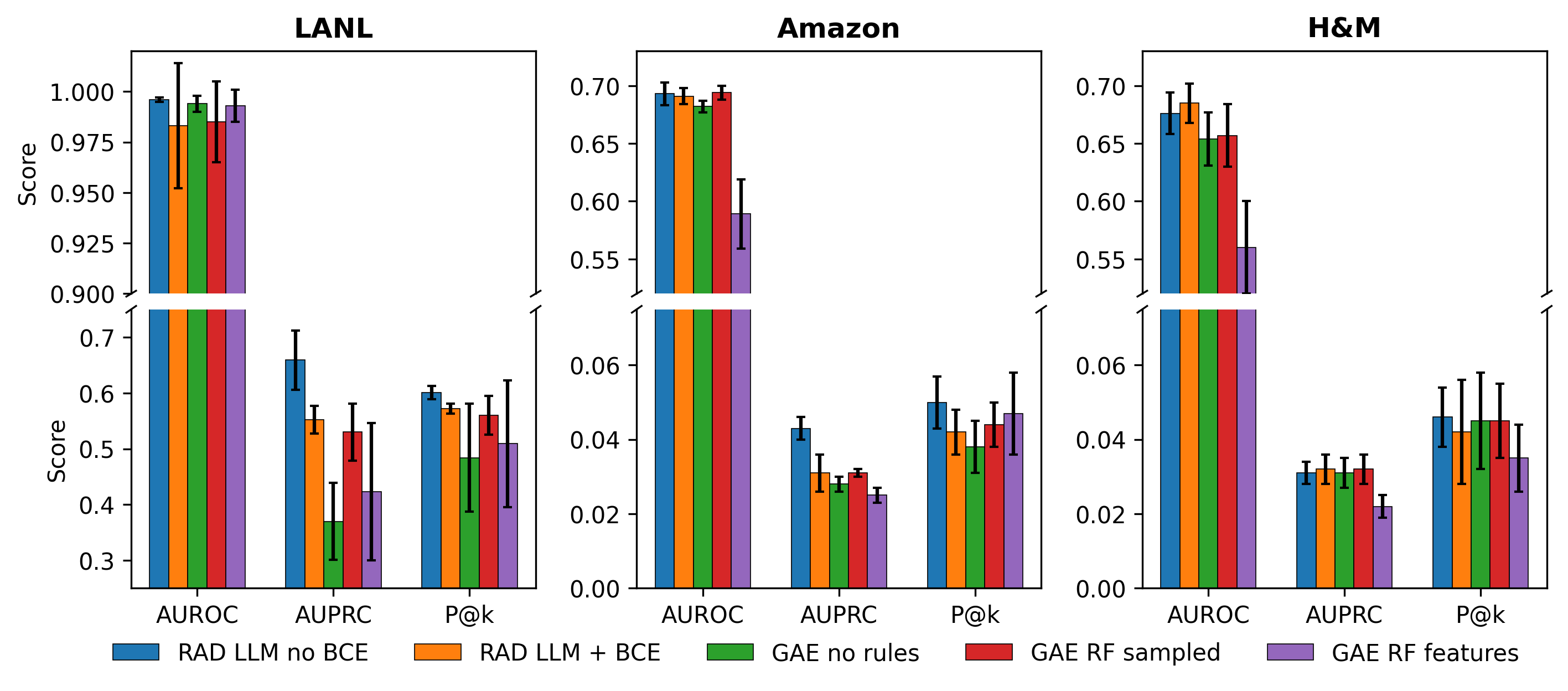}
    \Description{Rule-construction and structure-reconstruction ablation results across datasets over ten seeds. Bars show mean AUROC, AUPRC, and precision at k with standard-deviation error bars.}
    \caption{Rule-construction and structure-reconstruction ablations across datasets over 10 seeds. Bars show means and error bars show standard deviations. The y-axis is broken to separate AUROC from smaller AUPRC and P@k values.}
    \vspace{1ex}
    \label{fig:ablation}
\end{figure}

\vspace{0.5ex}
\subsection{RQ3: Effect of Structure Reconstruction}
\label{sec:rq3-results}

The effect of explicit graph structure reconstruction is mixed and dataset-dependent. In Table~\ref{tab:main-results}, the no-BCE RAD variant is strongest overall: it achieves higher average AUROC, AUPRC, and precision@k ranks than the BCE variant. On LANL, disabling BCE gives substantially stronger AUROC, AUPRC, and precision@k, suggesting that edge reconstruction may overemphasize artifacts of the sampled cybersecurity subgraphs. On Amazon, BCE and no-BCE perform similarly, with BCE slightly higher on AUROC and precision@k but essentially tied on AUPRC. On H\&M, the no-BCE variant is stronger across all three metrics.

\begin{figure*}[!t]
    \centering

    \begin{subfigure}[t]{0.45\textwidth}
        \centering
        \hspace{5ex}
        \includegraphics[width=0.75\linewidth]{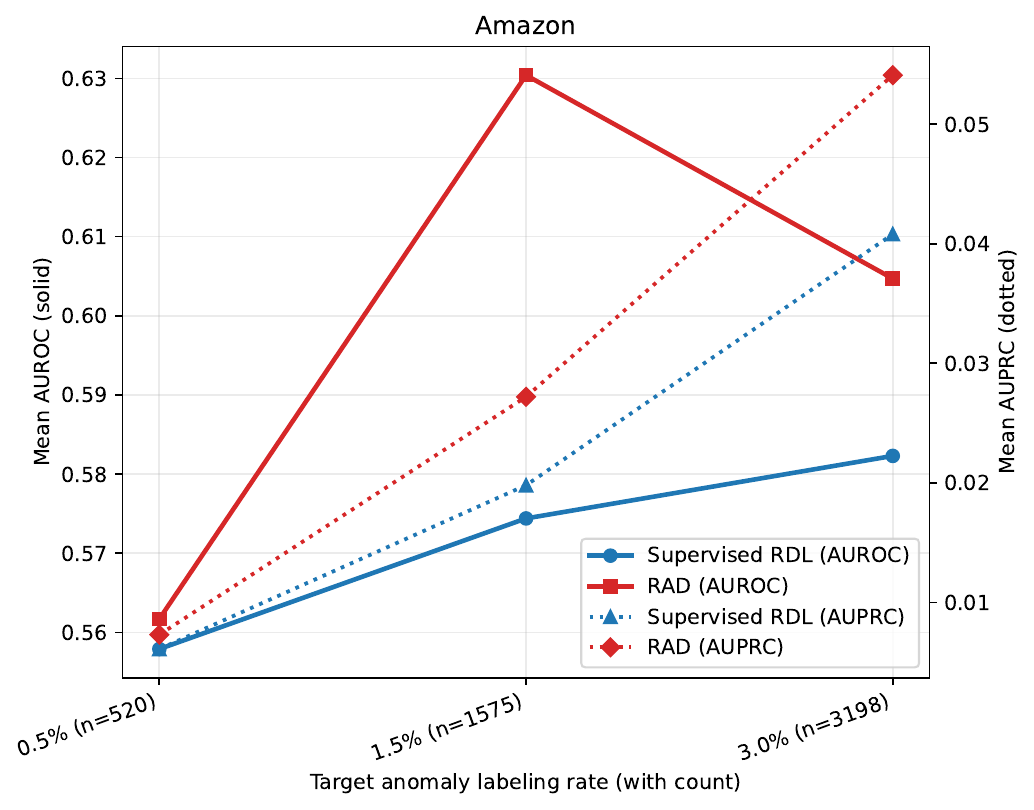}
        \caption{Amazon}
        \label{fig:prev-amazon}
    \end{subfigure}
    \hfill
    \begin{subfigure}[t]{0.45\textwidth}
        \centering
        \includegraphics[width=0.75\linewidth]{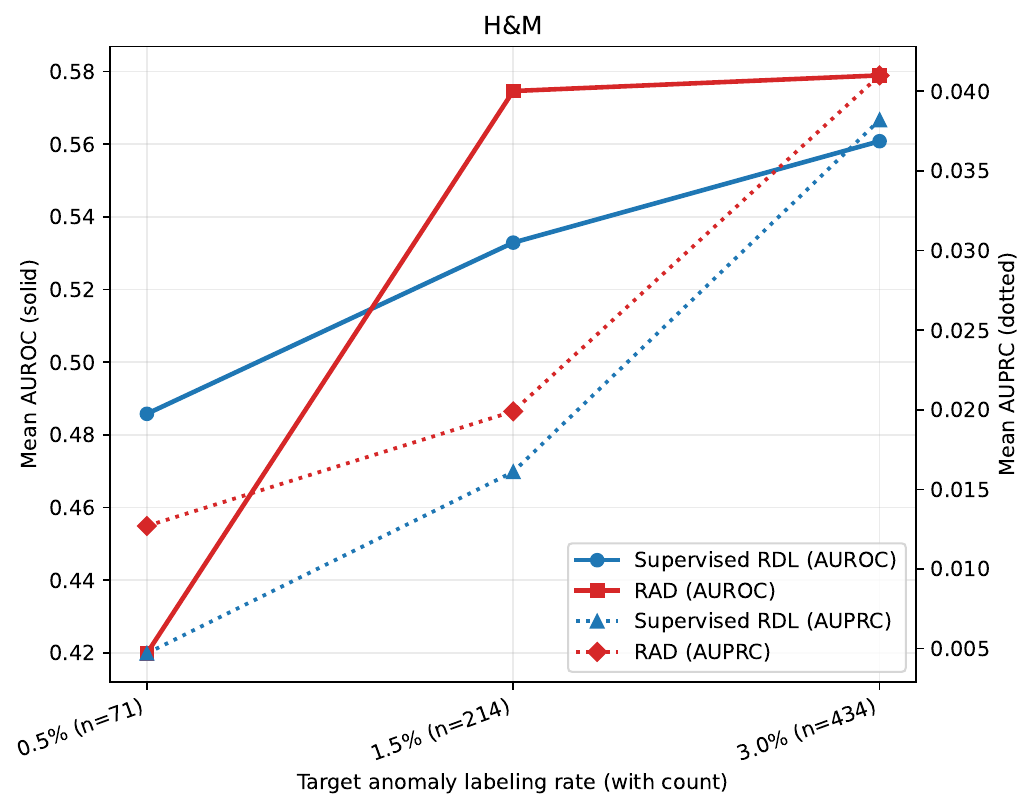}
                \hspace{5ex}
        \caption{H\&M}
        \label{fig:prev-hm}
    \end{subfigure}

    \Description{Prevalence sensitivity results for Amazon and H\&M comparing RAD and supervised RDL at 0.5, 1.5, and 3.0 percent anomaly prevalence.}
    \vspace{-2ex}
    \caption{Label-prevalence sensitivity for unexpected-churn detection. AUROC uses the left axis and AUPRC the right axis for target prevalences of 0.5\%, 1.5\%, and 3.0\%; counts show retained positives.}
    \vspace{0.5ex}
    \label{fig:prev}
\end{figure*}

These results suggest that edge reconstruction should be treated as an optional auxiliary objective rather than a uniformly beneficial component. It may help when anomalousness is expressed through relational topology, but it can also distract the model from anomaly ranking when the sampled graph structure is noisy or only indirectly related to the anomaly label.

\vspace{1ex}
\subsection{RQ4: Effect of LLM-Assisted Rule Refinement}
\label{sec:rq4-results}

LLM-assisted rule refinement provides a useful but not uniformly dominant improvement over random-forest-derived rules alone. The largest benefits appear on LANL, where refined rules improve anomaly retrieval compared with less compact rule representations. On Amazon and H\&M, refined rules remain competitive with random-forest sampled rules, suggesting that the LLM stage should be viewed primarily as a rule-filtering and rule-compaction step rather than as a replacement for random-forest rule mining.

Importantly, the LLM does not assign labels or produce anomaly scores. Its role is limited to proposing candidate predicates that are subsequently validated, filtered, and scored using the same training/validation rule-selection procedure described in Section~\ref{sec:rule-mining}. This keeps the anomaly detection pipeline grounded in the observed relational data and prevents test-label leakage.

\subsection{RQ5: Sensitivity to Unexpected-Churn Prevalence}
\label{sec:prevalence-results}

Figure~\ref{fig:prev} evaluates whether the unexpected-churn results depend on the target anomaly prevalence used in the label filter. Across 0.5\%, 1.5\%, and 3.0\% prevalence settings, the main performance trends remain broadly stable. On Amazon, RAD achieves its strongest AUROC at the default 1.5\% setting and its strongest AUPRC at 3.0\%, while supervised RDL improves more gradually as additional positives are retained. On H\&M, increasing prevalence generally improves performance, and RAD remains competitive and often stronger than supervised RDL across the sweep.

These results support using 1.5\% as the default benchmark configuration while showing that the conclusions are not driven by a single arbitrary prevalence threshold. The prevalence sweep also shows how performance changes as the anomaly definition becomes more permissive, while the relative comparison between RAD and the supervised relational baseline remains stable enough to support the benchmark design.

\subsection{Limitations}

RAD's rule mining is bounded by the auxiliary flattened view and requires labeled data for both the random-forest stage and support/confidence/lift ranking, so it does not extend as-is to unlabeled settings. Amazon and H\&M results are reported on a held-out partition of the validation split rather than a leaderboard test set, since RelBench withholds true test labels for these tasks. The unexpected-churn label itself is a derived construction; Section~\ref{sec:prevalence-results}'s prevalence sweep shows conclusions are stable across filter thresholds but the filter formula remains a design choice. LANL evaluation uses sampled subgraphs around anchor events rather than the full authentication log, and several tabular/relational baselines show substantial seed variance on LANL.

\section{Conclusion}
\label{sec:conclusion}

This paper introduced RAD, a rule-augmented framework for anomaly detection in relational databases. RAD is motivated by the observation that many anomalies are not isolated row-level outliers, but deviations whose meaning depends on relationships among entities, events, and historical behavior. Flattening relational data into a single feature matrix can obscure these dependencies, while relational representation learning alone does not provide explicit behavioral evidence or an anomaly-specific scoring objective.

RAD combines heterogeneous graph representation learning, mined symbolic rule features, reconstruction-based scoring, and ranking-based anomaly supervision. The relational graph preserves database structure, while the rule-mining component provides compact behavioral predicates derived from flattened summaries. By injecting rule features into target nodes before message passing, RAD allows symbolic evidence to influence learned relational representations rather than only post-hoc anomaly scores.

Across LANL cybersecurity event detection, Amazon review churn, and H\&M purchase churn, RAD improves anomaly retrieval under class imbalance, especially on AUPRC. The results suggest that relational anomaly detection is best treated as a joint problem of representation, scoring, and behavioral evidence. 

Several promising directions remain for future work. Scaling relational anomaly detection to larger databases and longer temporal histories will require more efficient graph construction and inference, consistent with recent work identifying scalability and temporal modeling as key challenges for relational deep learning \cite{dwivedi2025relational, lachi2025boosting}. RAD also could be extended toward unsupervised rule discovery and rules defined directly over complex multi-hop relational structure rather than flattened summaries. Finally, richer relational architectures \cite{chen2025relgnn} could provide even stronger backbones, while symbolic rules could be further leveraged to provide interpretable explanations for detected anomalies.

\vspace{-0.5ex}
\section*{Acknowledgments}

This work was supported by the National Security Agency (NSA) under grant number H98230-23-C-0279, and by the National Science Foundation (NSF) under grant numbers IIS-2239881, ECCS-2325417, IIS-2524380, and DGE-2622415.

\FloatBarrier

\vspace{-0.5ex}
\section*{GenAI Usage Disclosure}
Generative AI tools were used to assist with code development and manuscript writing. All experimental design, implementation choices, results, analysis, and final manuscript decisions were reviewed by the authors.

\balance 
\bibliographystyle{ACM-Reference-Format}
\bibliography{src/references}
\end{document}